\documentclass[	DIV=calc,%
							paper=a4,%
							fontsize=11pt,%
							twocolumn]{scrartcl}	 					

\usepackage{wrapfig,lipsum, tikz}													
\usepackage[hyperfootnotes=false]{hyperref}
\usepackage[english]{babel}										
\usepackage[protrusion=true,expansion=true]{microtype}				
\usepackage{amsmath,amsfonts,amsthm}					
\usepackage[hang, small,labelfont=bf,up,textfont=it,up]{caption}	
\usepackage{epstopdf}												
\usepackage{subfig}													
\usepackage{booktabs}												
\usepackage{fix-cm}													
\usepackage{natbib}
\usepackage{authblk}
\usepackage{graphicx}
\usepackage{caption}
\usepackage{afterpage}
\usepackage{tabularx}
\usepackage{newfloat}
\DeclareFloatingEnvironment[name={Supplementary Figure}]{suppfigure}
\usepackage{atbegshi}

\usepackage{sectsty}													
\allsectionsfont{
	\usefont{OT1}{phv}{b}{n}
	}

\sectionfont{
	\usefont{OT1}{phv}{b}{n}
	}

\usepackage{fancyhdr}												
\usepackage{lastpage}	

\usepackage{lettrine}

\usepackage{titling}															

\pretitle{\vspace{-30pt} \begin{center} 
				\fontsize{23}{23} \usefont{OT1}{phv}{b}{n} 
    \selectfont 
				}

\title{
NeuReg: Domain-invariant 3D Image Registration on Human and Mouse Brains
}

\posttitle{
\par\end{center}\vskip 0.1em
}

\author{Taha Razzaq, Asim Iqbal\thanks{Corresponding author: Asim Iqbal (asim@tibbtech.com)}\\
{\small Tibbling Technologies}\\
{\small Redmond, WA}
}

\date{\vspace{-0.7em}\fontsize{9}{11}\selectfont }

\begin{document}
\maketitle
\begin{tikzpicture}[overlay, remember picture]
  \node[xshift=-2.5cm,yshift=-1cm] at (current page.north east) {\includegraphics[width=0.4\textwidth]{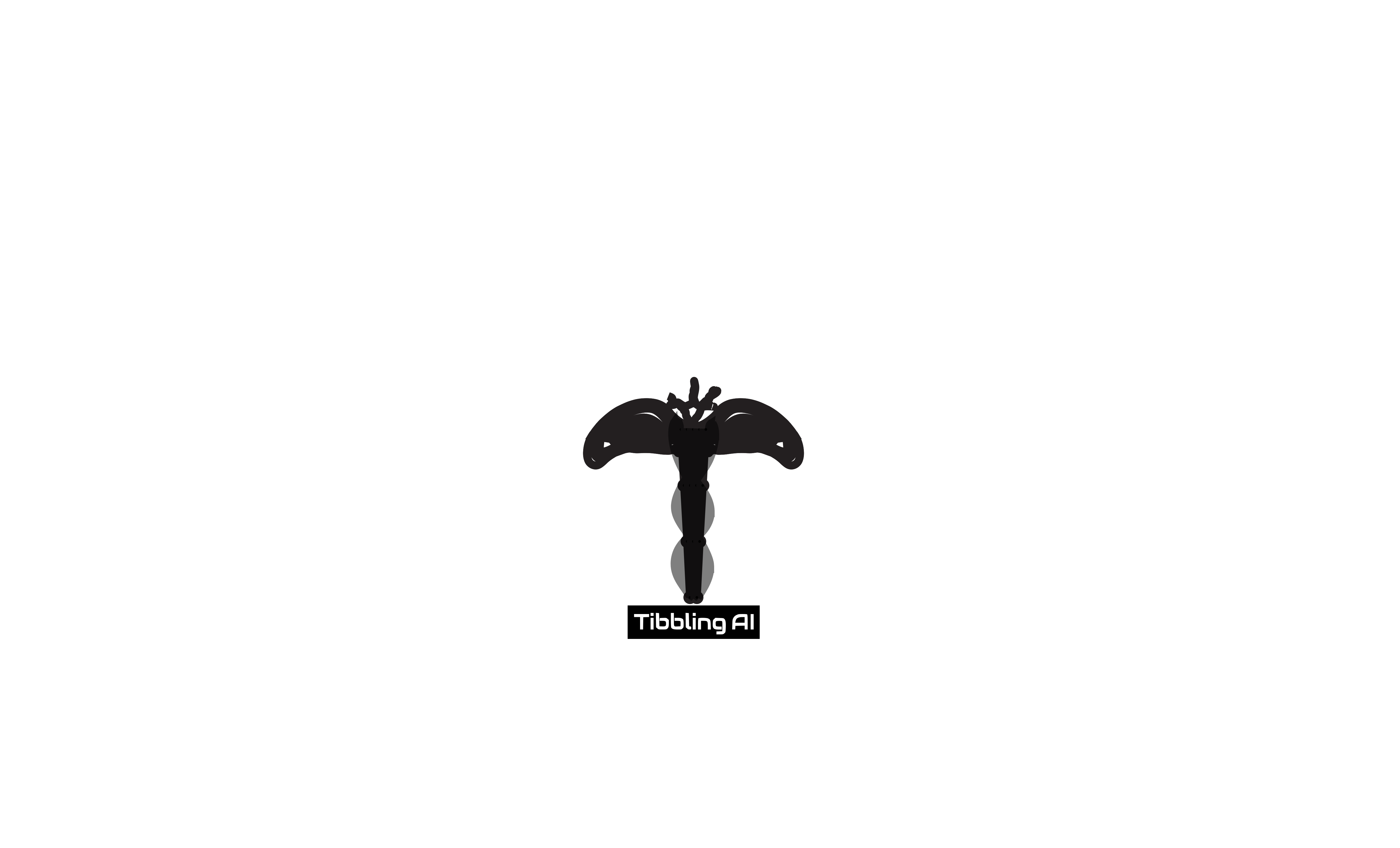}};
\end{tikzpicture}
\thispagestyle{fancy} 			
\noindent
\noindent
\hspace{-2.2em}
\textbf{Medical brain imaging relies heavily on image registration to accurately curate structural boundaries of brain features for various healthcare applications. Deep learning models have shown remarkable performance in image registration in recent years. Still, they often struggle to handle the diversity of 3D brain volumes, challenged by their structural and contrastive variations and their imaging domains. In this work, we present \textbf{\textit{NeuReg}}, a Neuro-inspired 3D image registration architecture with the feature of domain invariance. \textit{NeuReg} generates domain-agnostic representations of imaging features and incorporates a shifting window-based Swin Transformer block as the encoder. This enables our model to capture the variations across brain imaging modalities and species. We demonstrate a new benchmark in multi-domain publicly available datasets comprising human and mouse 3D brain volumes. Extensive experiments reveal that our model (\textit{NeuReg}) outperforms the existing baseline deep learning-based image registration models and provides a high-performance boost on cross-domain datasets, where models are trained on ``source-only'' domain and tested on completely ``unseen'' target domains. Our work establishes a new state-of-the-art for domain-agnostic 3D brain image registration, underpinned by Neuro-inspired Transformer-based architecture.}

\section*{Introduction}
Image registration is an integral part of medical imaging analysis which by learning spatial deformation is able to reveal significant differences between a pair of images or volumes. Traditional image registration approaches including ANTS (\cite{ants}), FFD (\cite{ffd}) and ADDMM (\cite{admm}) have achieved considerable performance and hence have been employed in various registration-based tools and techniques in real-world medical imaging applications, however, their time-consuming nature proves to be a bottleneck.
\\
\\
Recently, Deep Neural Networks (DNNs) have shown exponential success in computer vision tasks including object detection (\cite{obj_det}) and image classification (\cite{chen2021crossvit}), hence they are actively being used to replace the traditional image segmentation (\cite{iqbal2019developing}) and 3D registration models (\cite{c2fvit}, \cite{transmorph}), \cite{xmorpher}, \cite{mahmood2020exploring}, \cite{mahmood20233d}, \cite{mahmood2024multimodal}, \cite{mahmood2024neuroatlas}). Although, Convolutional Neural Networks (CNNs) efficiently encode and decode the visual differences between the input image pairs and provide an effective image registration pipeline (\cite{Huang_2021}), (\cite{label_driven}), there still exists a lot of limitations in these architectures in practice (\cite{shen2019networks}), hence efforts are being directed towards designing scalable deep learning architectures. In this direction, transformer-based deformable image registration models (\cite{transmorph}), (\cite{xmorpher}) have achieved state-of-the-art (SOTA) performance on different imaging datasets. However, they require extensive training which is resource- and time-consuming. To mitigate the extensive training, FourierNet (\cite{fouriernet}), proposed a band-limited deformation which converts the deformation field to a Fourier domain representation, and a model-driven decoder is used. We propose a novel architecture that integrates a Swin Transformer (\cite{swin}) encoder with a model-driven decoder, enabling high performance in the 3D registration task while minimizing the computational resources required for model training. 
\\
Furthermore, a significant drawback of using DNNs and transformer-based architectures is their dependency on the training data domain. In case of a covariate shift between the training and testing data, these models experience performance deterioration. Recent models like SynthMorph (\cite{synthmorph}) mitigate the dependency on the training data domain by drawing training samples from a random distribution. While this approach leads to a contrast invariant model yielding significant results, it requires prolonged training and fails to capture scale and structure variation in complex datasets.
\\
\\
We design a neuro-inspired pre-processing layer in our architecture that generates a domain-agnostic representation of the 3D brain volumes. To combine this domain-agnostic representation feature with the Swin Transformer encoder and model-driven decoder, we propose, \textit{NeuReg}, a domain-agnostic transformer-based model architecture which is inspired by the mammalian visual system, and ensures that our 3D image registration model performs well on 'unseen' target domains for human and mouse datasets.
\section*{Results and Discussion}
\textbf{Performance on cross-domain human datasets:} 
\\
\noindent
To quantify the performance of our model on human brain imaging, we consider two open-sourced, commonly used Magnetic Resonance Imaging (MRI) datasets. iSeg-2017 dataset contains multi-modal MRI data covering T1-weighted and T2-weighted imaging domains. While T1-weighted MRI enhances the intensity of the fatty tissue and suppresses the signal of the Cerebrospinal Fluid (CSF) in the brain, T2-weighted MRI is completely opposite in terms of intensity since it enhances the CSF signal instead. Models trained on either domain (T1 or T2) are expected to perform well on MRI scans generated from the same domain but may perform poorly in case of a domain shift however a domain-agnostic framework should generalize and have high performance even across domains. To gauge the extent to which our model architecture generalizes over different domains, we compare its performance with the baseline models for both T1 and T2 domains in the iSeg-2017 dataset. We report the DICE and SSIM score for the iSeg-2017 dataset in {\hyperref[tab:iseg]{\textbf{Table 1}}} along with qualitatively comparing our model's results with FourierNet and SynthMorph, as shown in {\hyperref[fig:fig2]{\textbf{Figure 2}}}. Both qualitative and quantitative results demonstrate that our model outperforms the existing 3D registration models. Although the baseline architectures have benchmarking performance for 3D registration on large open-sourced single-domain human MRI datasets, they are unable to generalize over different domains. Since our model, while creating a domain-agnostic representation of the data, mitigates the intensity shift present in the T1 and T2 scans, it cannot only outperform SOTA models but also achieve significantly high SSIM scores.
\\
\\
{\hyperref[fig:fig2]{\textbf{Figure 2}}} qualitatively compares our model with SynthMorph and FourierNet while highlighting the CSF region. The zoomed in snippets of the normal brain ({\hyperref[fig:fig2]{\textbf{Figure 2}}}, $1^{st}$ and $3^{rd}$ columns) clearly show the intensity difference between T1 and T2. While T1 has a low-intensity signal for CSF, the region has a high intensity in the case of T2. Moreover, the snippets showing our version of the fixed and moving samples ({\hyperref[fig:fig2]{\textbf{Figure 2}}}, $2^{nd}$ and $4^{th}$ columns), clearly highlight that our model removes the intensity shift, and CSF in domain generalized T1 and T2 has attained similar intensity profile. The registration results show that while SynthMorph has an overall decent performance,  it falters in specific regions leading to a marginally low score, as compared to our models. In the case of FourierNet, although the model tries to align the moving segmentation to the ground truth it fails in regions where intensity shift is encountered, including CSF. Our models, on the other hand, are able to come close to the actual segmentation ground-truth, and hence achieve high SSIM and DICE scores. This trend is consistently observed in cases where the model is trained on T1 and tested on T2 and vice versa.
\\
\\
Since iSeg-2017 is a relatively small dataset, in order to show the utility of our model across a large dataset, we test our framework on larger and more complex OASIS-3 brain imaging data. We focus on T1-weighted and T2-weighted MRI scans and report results on both domains. Since SynthMorph and FourierNet are the current state-of-the-art in 3D registration, we compare the performance of our model with them for the OASIS-3 dataset.
\\
\\
In {\hyperref[tab:oasis]{\textbf{Table 2}}} we report the SSIM score computed between the fixed and aligned moving brains as the metric to compare the performance of our model with SynthMorph and FourierNet. We observe that both SynthMorph and FourierNet have a very low performance on data sampled from unseen domains which indicates that the model is unable to capture the varying complexity in different domains and generalizes poorly. Our model architecture, however, outperforms them by a large margin, indicating its ability to generalize across different domains. OASIS-3 dataset not only spans different domains but also contains scans from participants at different stages of cognitive decline, hence leading to large structural variation. Since we do not differentiate between the patients' cognitive health, the dataset also encapsulates structural changes. This adds to the overall complexity of the dataset and hence it requires extensive training. Although, our model outperforms SynthMorph and FourierNet and can capture the dataset complexity, however, we anticipate that our model results can be further improved with extensive training. We show additional results and comparison of our model with SynthMorph and FourierNet in {\hyperref[fig:supp_fig1]{\textbf{Supplementary Figure 1}}}
\\
\\
\noindent
\textbf{Performance on cross-domain mouse dataset:}
\noindent
To show the generalizability of our approach across species and complex imaging domains, we also repeat our experiments on the DevCCF dataset. Each model is trained on a single domain and tested on the remaining $5$ domains, for which the average SSIM and DICE score is reported in {\hyperref[tab:devccf]{\textbf{Table 3}}}. 
\\
\\
The qualitative comparison between our best performing model and FourierNet is also shown in {\hyperref[fig:fig3]{\textbf{Figure 3}}}. The first row in {\hyperref[fig:fig3]{\textbf{Figure 3}}} shows a clear intensity-based shift in the original samples from all the domains along with their scale variations.
\\
\\
Since DevCCF encapsulates different developmental (Postnatal day 4, 14, and 56) stages in the mice brain, the dataset has significant variation in terms of scale and structure. While the domains vary in intensity and scale factors, they also have structural dissimilarities since the dataset spans different postnatal mouse ages. Since the brain undergoes delineation during development and each brain region grows in a non-uniform fashion, the dataset contains significant structural variation. A domain-agnostic registration model trained on DevCCF is expected to overcome scale as well as structural variation in 3D brain volumes and perform well across ages and domains. Due to the excessive structural dissimilarities in this dataset, our model trained with MSE as the image similarity loss performs the best since it is able to capture the existing structural differences between the input images (fixed and moving) during training. Moreover, since we aim to evaluate the models' capability to register across scale and structure variation, we use SSIM as the evaluation metric as well. Both {\hyperref[tab:devccf]{\textbf{Table 3}}} and {\hyperref[fig:fig3]{\textbf{Figure 3}}} show that our model performs better than the baselines and is able to capture the variation and complexity of the dataset in an apt manner. Additional results on different postnatal mouse ages is shown in {\hyperref[fig:supp_fig7]{\textbf{Supplementary Figure 2}}}. We show that our model outperforms FourierNet on all ages. 
\\
\\
Through qualitative and quantitative results, we show that our model (NeuReg) outperforms existing registration frameworks across modalities and species. We report benchmarking performance for publicly available human and mouse datasets and show our model's capability to generalize across domains.


\section*{Conclusion}
In this study, we propose NeuReg, a neuro-inspired domain agnostic transformer-based architecture for 3D image registration and demonstrate its benchmarking performance on cross-modal publicly available human and mouse brain 3D imaging datasets. Unlike the existing baselines and SOTA models for registration, our approach generalizes over unseen domains and performs exceptionally well on multi-domain datasets. Our domain generalization pre-processing layer in the proposed NeuReg architecture generates an invariant representation of the imaging volumes that captures the key common features present in cross-modal datasets. We demonstrate the effectiveness of our approach on commonly used human brain cross-modal datasets, for instance, training on T1 MRI modality and testing on T2 MRI modality (using iSeg-2017 and OASIS-3 datasets) and vice versa. We further extend this on the mouse brain 3D imaging dataset and demonstrate the benchmarking performance across all six cross-domains (by training on a single source domain and testing on the remaining 5 domains for each category). Our Swin-transformer-based architecture not only captures the intensity-based shifts present in the cross-modal datasets but also captures the scale and non-linear delineations present in longitudinal datasets (by training on a single developmental age and testing on unseen developmental ages with varying domains). This shows the promising performance of our proposed architecture on multiple fronts i.e. domain-shift due to imaging as well as age. In the future, we aim to extend NeuReg to handle more complex domains and even diverse datasets as well as apply it to explore neurodevelopmental alternations in healthy and non-healthy brains.

\section*{Methods}
\subsection*{NeuReg architecture}
\begin{figure*}[!htb]
    \centering
    \includegraphics[width=\textwidth]{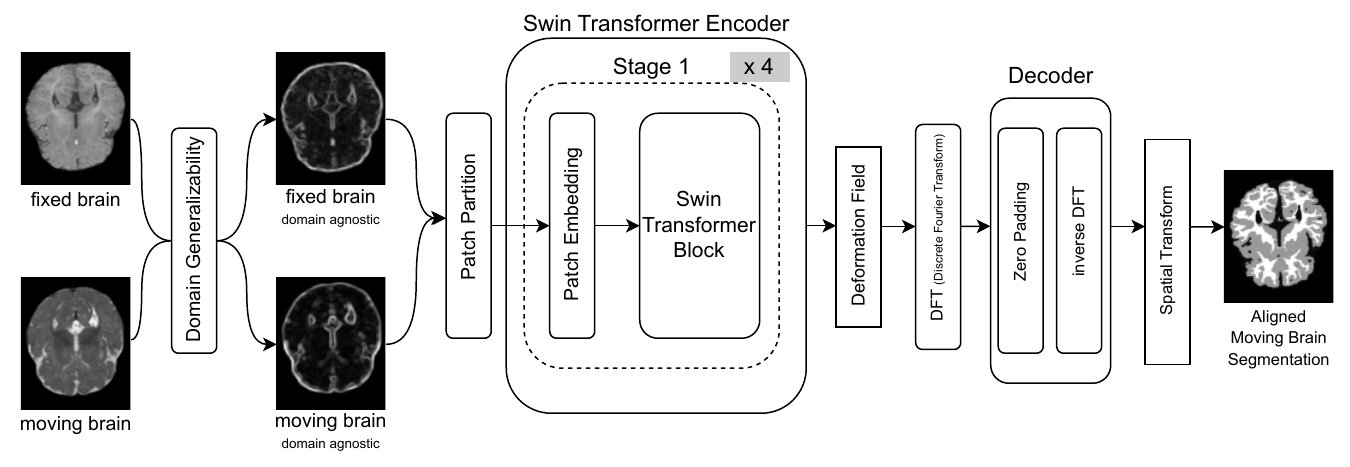}
    \caption{Block diagram for the proposed methodology. Input (fixed and moving) samples are passed to the domain generalised layer followed by patch partition which splits the input samples into patches which are fed to the swin transformer encoder referred to as Stage 1. Our architecture consists of four such repeated stages after which the the Swin transformer encoder generates the deformation field. A Discrete Fourier Transform (DFT) is applied to convert the deformation field to a fourier domain, which is then passed through the model-driven decoder. The moving brain segmentation is warped with the deformation field via a spatial transform to generate the final output.}
    \label{fig:fig1}
\end{figure*}

\begin{figure*}[!htb]
    \centering
    \includegraphics[width=\textwidth]{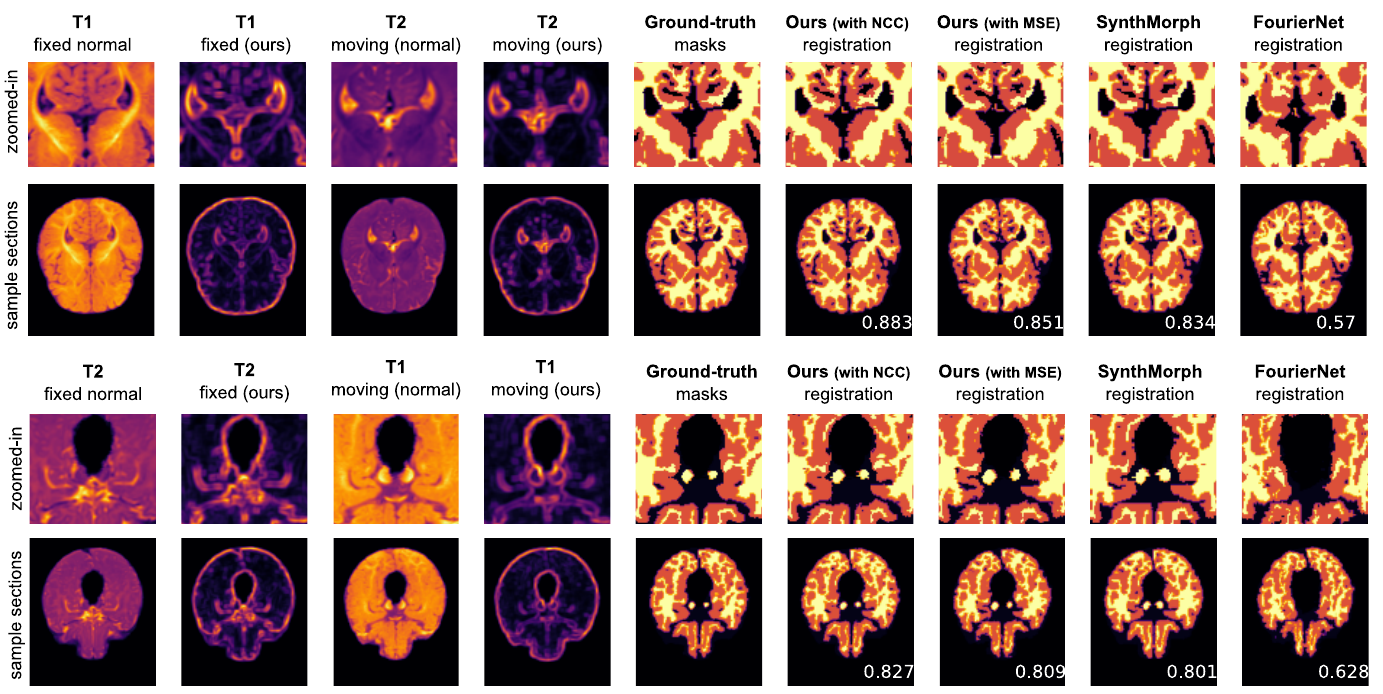}
    \caption{Qualitative results on iSeg-2017 dataset. Top rows show the zoomed-in version of the normal and our model's data representation, followed by the ground truth segmentation, aligned moving segmentation generated by our model with NCC and MSE and the aligned moving segmentation generated by SynthMorph and FourierNet. Bottom rows show the corresponding samples with the DICE score mentioned underneath.}
    \label{fig:fig2}
\end{figure*}

\begin{figure*}[!htb]
    \centering
    \includegraphics[width=\textwidth]{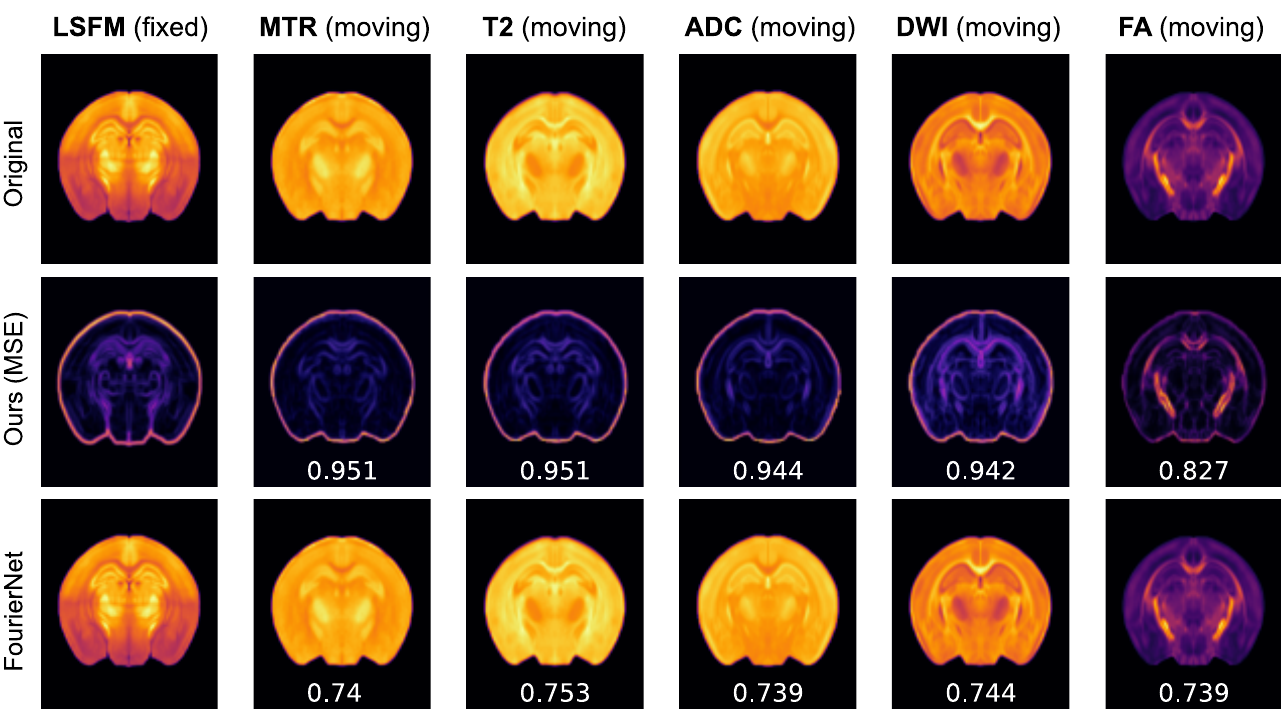}
    \caption{Qualitative results on DevCCF dataset. The original samples from all the domains are shown (top row) followed by the output of our top-performing model (middle row) and FourierNet output (bottom row) with the corresponding SSIM mentioned underneath.}
    \label{fig:fig3}
\end{figure*}

\noindent
We follow the conventional paradigm of image registration where a moving brain $I_m$ is mapped to the fixed brain, $I_f$. We propose a novel domain-agnostic transformer-based architecture, summarized in {\hyperref[fig:fig1]{\textbf{Figure 1}}}, which consists of a domain generalized pre-processing layer, Swin transformer encoder, and a model-driven decoder. 
\\
\\
\noindent
\textbf{Domain generalization layer:}
\noindent
We take inspiration from the feature processing procedure by the excitatory and inhibitory neurons in the mammalian visual cortex and design a domain-agnostic representation of image. In 3D registration, we process the input 3D volume \(I \in \mathbb{R}^{W \times H \times D}\), where \(W\), \(H\), and \(D\) represent its width, height, and depth, respectively. Our goal is to form a domain-independent representation, denoted as \(I_{(n)}\). The image \(I\) is segmented into patches of size \(x\), represented as \(T = \{t_{1}, t_{2}, \ldots, t_{n}\}\), where each patch \(t_i\) encircles a pixel \(m\) at coordinates $(i, j, k)$. For each patch, its mean \(\mu_{t_i}\) and standard deviation \(\sigma_{t_i}\) are calculated to construct \(I_{n}\):
\\
\\
    \begin{equation}
        \mu_{t_i} = \frac{1}{x^2} \sum_{i,j,k \in t_i} m_{ijk}
    \end{equation}
    \begin{equation}
        \sigma_{t_i} = \left( \frac{1}{x^2} \sum_{i,j,k \in t_i} (m_{ijk} - \mu_{t_i})^2 \right)^{1/2}
    \end{equation}
The corresponding domain-generalized representation, \(I_{n}\) is derived by updating pixel values within each patch according to the mean and standard deviation of the patch, as governed by the following equations
    \begin{equation}
        z = \max \{\sigma_{t_1}, \sigma_{t_2}, \ldots, \sigma_{t_n}\}
    \end{equation}
    \begin{equation}
        I_{(n)} = \sigma_{t_i} / z
    \end{equation}
Existing pre-processing contrast normalization techniques, including local contrast normalization (LCN) (\cite{lcn_paper}) and local response normalization (LRN) (\cite{lrp}), often result in information loss. These methods focus on enhancing local contrast within an image, which can lead to the suppression or enhancement of specific pixels, thus losing important pixel-level information (\cite{lcn}). Similar issues arise with edge detection filters such as Laplacian (\cite{laplacian}) and Sobel (\cite{Sobel}) filters, which primarily detect edge maps and tend to overlook low-intensity pixel information.
\\
\\
In contrast, our technique preserves pixel-level information while ensuring normalization across the entire image without selectively enhancing or suppressing any pixels. This results in a consistent and comprehensive contrast normalization. Additionally, our domain generalization technique can serve as a pre-processing layer atop any existing 3D registration model, enhancing performance for multi-domain registration tasks ({\hyperref[tab:dg]{\textbf{Supplementary Table 1}}}).
\\
\\
The domain agnostic representation of $I_m$ and $I_f$ is fed to the Swin Transformer for registration. 
\\
\\
\noindent
\textbf{Swin transformer encoder:}
\noindent
For our encoder block, we use rectangular parallelepiped windows, a variation of the original Swin transformer, to accommodate the varying shapes of the 3D volumes ($M_x, M_y, M_z$). 
The global and local differences between the domain invariant image input pairs ($I_m$ and $I_f$) are successfully captured by the self-attention mechanism of the Swin transformer and are used to produce a deformation field ($\phi$). We apply general Discrete Fourier transform (DFT) on the deformation field and convert it from a spatial domain to a low pass Fourier domain. 
\\
\\
\noindent
\textbf{Model-driven decoder:}
\noindent
The decoder used in our framework comprises a zero padding layer followed by an inverse DFT (iDFT). The low dimensional deformation fields produced by the encoder are passed through a zero padding layer and a standard inverse Discrete Fourier Transform is applied to restore the original dimensions of the deformation field ($\phi$). Since both zero padding and iDFT are differentiable, our decoder does not require training and is able to manipulate the encoder's output through knowledge-based learning only. The resultant deformation fields are warped with $I_m$ through spatial transform to produce the aligned moving brain $I_w$. 
\\
\begin{equation}
    I_w = I_m \circ \phi
\end{equation}

\subsection*{Loss functions}
\noindent
The loss function used in our model training, similar to the traditional frameworks, consists of an image similarity loss which is computed between the fixed and the aligned moving image along with a regularizer which ensures the smoothness of the generated deformation fields. 
\begin{equation}
    \mathcal{L}\left(I_f, I_m, \phi\right)=\mathcal{L}_{\text {sim }}\left(I_f, I_m, \phi\right)+\lambda \mathcal{R}(\phi)
\end{equation}
where $\mathcal{L}_{\text {sim}}$ represents the image similarity measure and $\mathcal{R}$ is the deformation field regularizer. $\lambda$ is the hyper-parameter for balancing both terms.

\section*{Experiments}
\subsection*{Datasets and pre-processing}
\noindent
We set up our experiments to analyze the performance of our domain-agnostic transformer-based architecture for 3D image registration. We train the existing state-of-the-art (\cite{fouriernet},\cite{transmorph},\cite{voxelmorph}) and our model on a specific domain and test their performance on remaining, completely unseen domains. To gauge SynthMorph's performance, we utilized its publicly available brain image registration model checkpoint and conducted experiments on human and mouse brain 3D volumes.
\\
\\
\noindent
\textbf{Human datasets:}
\noindent
For our experiments we use two open-sourced Human datasets - iSeg-2017 dataset (\cite{iseg}) and OASIS-3 (\cite{oasis3}). iSeg-2017 dataset comprises of brain tissues from T1 and T2-weighted MRI imaging data collected from $6$-month-old infant brains. The brain scans are accompanied by their segmentation which partitions the brain into $3$ distinct regions - White matter (WM), Grey matter (GM), and Cerebrospinal Fluid (CSF). Since the iSeg-2017 dataset is publicly available and contains multi-domain brain imaging data, it perfectly aligns with our study. This dataset is commonly used by the community for 3D segmentation frameworks (\cite{iseg_work}, \cite{iseg_work_2},\cite{iseg_work_3}). 
\\
\\
Another publicly accessible human dataset, OASIS-3, is used in our experimentations. OASIS-3 is a retrospective compilation of data collected by WUSTL Knight ADRC over the course of $30$ years. The dataset contains MR sessions from normal participants along with individuals undergoing cognitive decline. The MRI imaging data spans T1-weighted, T2-weighted, FLAIR, ASL, SWI, time of flight, resting-state BOLD, and DTI domains. For our experimentation, we shortlisted T1-weighted and T2-weighted MRI scans and remained agnostic to the participants' cognitive health. OASIS-3 is also actively used for cognitive health prediction frameworks, especially for Alzheimer's disease (\cite{oasis_work},\cite{oasis_work_2}). Standard pre-processing of skull stripping and normalization was performed by the provider and all the brain scans were affinely registered using a common scan before being cropped to ($160 \times 192 \times 224$) dimensions.\\
\\
\noindent
\textbf{Mouse dataset:}
\noindent
In order to demonstrate the generalizability of our proposed framework across species, we also perform all the experiments on the DevCCF mouse imaging dataset (\cite{devccf}). DevCCF is an open-sourced 3D mouse brain imaging cross-domain dataset that contains multi-modal mouse brain volumes collected at $7$ different developmental timestamps. While this dataset also covers mouse embryonic days (E11.5, E13.5, E15.5, E18.5), we only focus on the Postnatal (P) days (P4, P14 and P56). All the domains present in DevCCF (LSFM, FA, DWI, ADC, MTR, and T2-weighted) are used in our experiments. We trained ours as well as baseline models on a single domain tested the performance on all the remaining unseen domains and repeated the same process for each domain.
\\
\\
We focus on inter-subject registration, which is a core and challenging problem due to spatial complexity. During training, we pair samples from different subjects within the same domain while for testing, we combine samples from different subjects and domains. All models were trained and tested on a consistent dataset. We trained our model for $500$ epochs with early stopping (patience $= 30$). The hyper-parameters were optimized and learning rate $lr = 5e^{-4}$ and $\lambda = 1.0$ yielded the best results. When MSE is used as the image similarity loss, $\lambda$ is set to $0.2$, yielding the best results across all models.

\subsection*{Baseline architectures}
\noindent
To gauge the performance of our proposed architecture, \textit{NeuReg}, we compare our results with the SOTA 3D registration model, FourierNet (\cite{fouriernet}). However, since FourierNet is not trained in a domain-agnostic manner, we also compare our model's results with SynthMorph (\cite{synthmorph}), which has SOTA performance on out-of-domain registration tasks. 
For the iSeg-2017 and DevCCF datasets ({\hyperref[tab:supp_devccf]{\textbf{Supplementary Table 2}}}), we also show comparison with additional baselines - Transmorph (\cite{transmorph}) and VoxelMorph (\cite{voxelmorph}).  
\\
\\
For comparison on the OASIS-3 dataset, FourierNet and SynthMorph are only used and we do not perform comparison with remaining baselines to minimize unnecessary GPU usage for environmental responsibility. We expect the results of OASIS-3 to align with the iSeg-2017 dataset given their similar nature.

\subsection*{Evaluation metrics}
\noindent
To quantify the performance of our proposed architecture, we use DICE (\cite{dice}) score and SSIM (\cite{ssim}) metrics, which are commonly used in 3D registration. For the iSeg-2017 and DevCCF datasets, we report the DICE score computed using the segmentation (labels) ground truth and the aligned moving segmentation, along with SSIM. As for OASIS-3, we report SSIM  between the fixed and aligned moving brains.
\\
\\
Since SSIM is a combination of comparison between luminance, contrast and structural similarity between the two volumes, it is a well-suited evaluation metric for our proposed architecture which manipulates the intensity and scale variations in the scans.


\bibliographystyle{unsrtnat}
\bibliography{main} 
\medskip

\newcolumntype{Y}{>{\centering\arraybackslash}X}

\begin{table*}[!htb]
\centering
\caption{Quantitative results on $iSeg-2017$ dataset}
\label{tab:iseg}
\scalebox{0.8}{ 
\small

\begin{tabularx}{1.25\linewidth}{l *{5}{Y} }
   \toprule
    & \multicolumn{2}{c}{$T1$ (train)$ \rightarrow$ $T2$ (test)} & \multicolumn{2}{c}{$T2$ (train) $\rightarrow$ $T1$ (test)}
    \\
    \cmidrule(l){2-3} \cmidrule(l){4-5}\\

Model & {DICE} & {SSIM} & DICE & SSIM \\
\midrule
VoxelMorph & $0.4765$ & $0.9064$ & $0.4729$ & $0.902$ \\
TransMorph & $0.473$ & $0.8671$ & $0.5495$ & $0.8832$ \\
Fourier Net & $0.5934$ & $0.8872$ &
$0.5991$ & $0.8881$\\
SynthMorph & $0.6571$ & $0.8947$ & $0.6605$ & $0.8942$ \\
Ours (MSE) & $0.6338$ & \textbf{0.9297} & $0.6382$ & \textbf{0.9307} \\
Ours (NCC) & \textbf{0.6767} & $0.9225$ & \textbf{0.6793} & $0.9257$ \\
\bottomrule
\end{tabularx}
} 
\end{table*}

    
\newcolumntype{Y}{>{\centering\arraybackslash}X}
\begin{table*}[!ht]
\centering
\caption{Quantitative results (SSIM score) on OASIS-3 dataset}
\label{tab:oasis}
\scalebox{0.8}{ 
\normalsize

\begin{tabularx}{1.25\linewidth}{l *{3}{Y} }
   \toprule
Model & $T1$ (train)$ \rightarrow$ $T2$  (test)  & $T2$ (train) $\rightarrow$ $T1$
 (test) \\
\midrule
Fourier Net & 0.1857 & 0.1309 \\
SynthMorph & 0.0157&
0.0126 \\
Ours (MSE) &0.6231 & \textbf{0.6568}
\\
Ours (NCC) & \textbf{0.6434} &
0.6214
\\ 
\bottomrule
\end{tabularx}
}
\end{table*}


\newcolumntype{Y}{>{\centering\arraybackslash}X}

\begin{table*}[!ht]
\centering

\caption{Quantitative results for DevCCF dataset. We train the models on each domain, and report performance on all the remaining (unseen) domains.}
\label{tab:devccf}

\scalebox{0.8}{ 
\small

\begin{tabularx}{1.25\linewidth}{@{\extracolsep{\fill}} l *{13}{Y} @{} }
   \toprule
    & \multicolumn{2}{c}{LSFM} & \multicolumn{2}{c}{MTR} & \multicolumn{2}{c}{T2} & \multicolumn{2}{c}{ADC} & \multicolumn{2}{c}{DWI} & \multicolumn{2}{c}{FA}
    \\
    \cmidrule(l){2-3} \cmidrule(l){4-5} \cmidrule(l){6-7}  \cmidrule(l){8-9} \cmidrule(l){10-11} \cmidrule(lr){12-13}
Model & DICE  &  SSIM  & DICE  &  SSIM  & DICE  &  SSIM  & DICE  &  SSIM  & DICE  &  SSIM  & DICE  &  SSIM  
 \\
\midrule
Fourier Net  &
 0.2346 & 0.7284 & 
0.1799 &0.7699 & 
0.1988 & 0.7316 & 
0.2069 & 0.7996 & 
0.2224 & 0.7281 & 
0.1474 & 0.8687\\
SynthMorph  &
0.0381 & 0.7357 & 
0.185 & 0.7817 & 
0.168 & 0.7603 & 
0.181 & \textbf{0.8361} & 
0.169 & 0.7626 & 
0.177 & 0.7782 
 \\
Ours (MSE) & \textbf{0.298} & \textbf{0.820} & 
\textbf{0.305} & \textbf{0.815} & 
\textbf{0.291} & \textbf{0.8095} & 
\textbf{0.311} & 0.831 & 
0.3004 & \textbf{0.8524} & 
\textbf{0.295}& \textbf{0.913}  \\

Ours (NCC) & 0.2961 &  0.7956 & 
0.2806 &  0.7909 & 
0.2887 & 0.7835 & 
0.2905 & 0.8037 & 
\textbf{0.3054} &0.8223 & 
0.2554 & 0.8901\\ 

\bottomrule
\end{tabularx}

}
\end{table*}


\clearpage

\clearpage

\clearpage
\clearpage
\renewcommand{\tablename}{Supplementary Table}
\setcounter{table}{0}


\newcolumntype{Y}{>{\centering\arraybackslash}X}

\begin{table*}[!htb]
\centering

\caption{Quantitative results (DICE score) on iSeg-2017 dataset with and without adding our domain generalization layer. For all models, our domain generalization layer leads to a performance boost.}
\label{tab:dg}

\scalebox{1.0}{ 
\small

\begin{tabularx}{\textwidth}{l *{5}{Y} }
   \toprule
    & \multicolumn{2}{c}{$T1$ $\rightarrow$ $T2$ } & \multicolumn{2}{c}{$T2$ $\rightarrow$ $T1 $} 
    \\
    \cmidrule(l){2-3} \cmidrule(l){4-5}
Model & {$w/o$ DG} & {with DG} & w/o DG & with DG \\
\midrule

Fourier Net & 0.5934 &  \textbf{0.6279} & 0.5991 & \textbf{0.615} \\

Ours (MSE) & 0.5875 & \textbf{0.6338} & 0.5826 &
\textbf{0.6382} 
\\
Ours (NCC) & 0.594 & \textbf{0.6445} & 0.5901 & \textbf{0.6483} 
\\ 

\bottomrule
\end{tabularx}

}

\end{table*}


\newcolumntype{Y}{>{\centering\arraybackslash}X}

\begin{table*}[!ht]
\centering

\caption{Quantitative results (SSIM score) for DevCCF dataset. We show benchmarking performance of our model compared to the baselines.}
\label{tab:supp_devccf}

\scalebox{0.8}{ 
\small

\begin{tabularx}{1.25\linewidth}{@{\extracolsep{\fill}} l *{7}{Y} @{} }
   \toprule
Model & LSFM  &  MTR  &  $T2$  & ADC  &  DWI & FA
 \\
\midrule
VoxelMorph & 0.7935 & 0.8056 & 0.78390 & 0.7088& 0.7036 & 0.7449\\
TransMorph & 0.7318  &0.7647 & 0.7585 & 0.7530 & 0.7558 & 0.7635\\

Ours (MSE) & \textbf{0.8205} & 
\textbf{0.8152} & 
\textbf{0.8095} & 
\textbf{0.8314} & 
\textbf{0.8524} & 
\textbf{0.9129}  \\

\bottomrule
\end{tabularx}

}
\end{table*}


\clearpage

\renewcommand{\figurename}{Supplementary Figure}
\setcounter{figure}{0}
\begin{figure*}[!htb]
    \centering

    \includegraphics[scale=0.8]{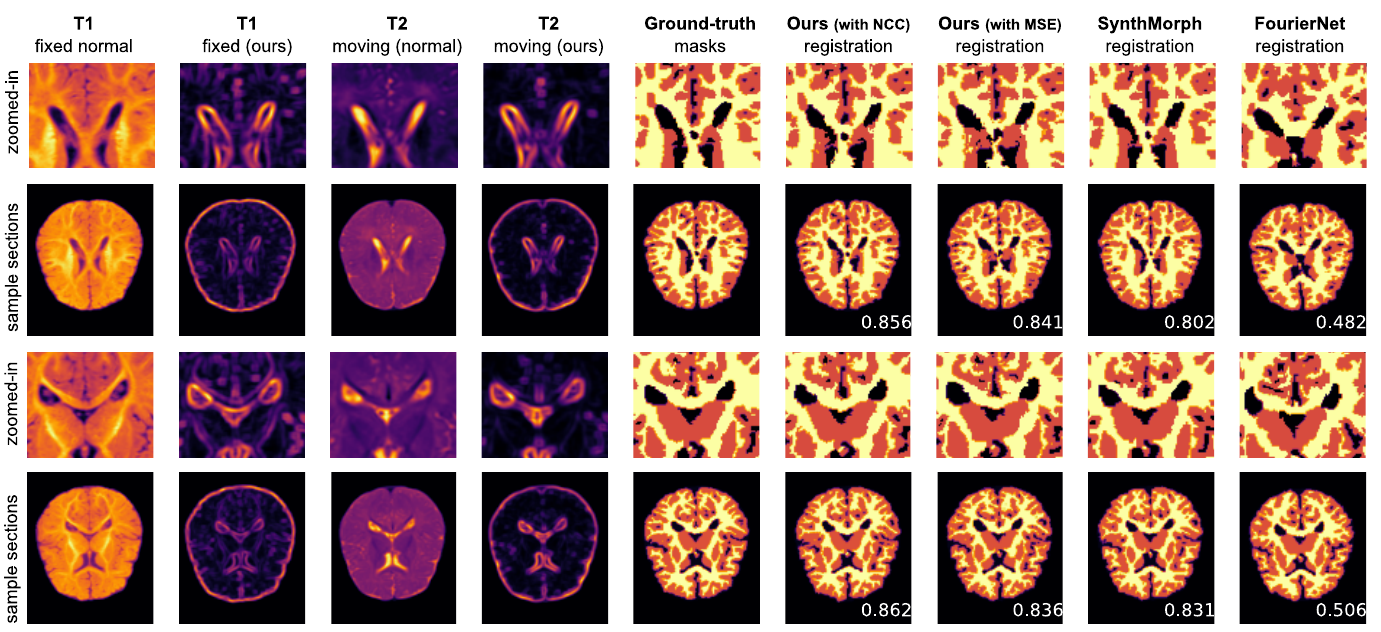}
   \caption{Qualitative results on iSeg-2017 dataset. Top rows show the zoomed-in version of the samples shown in the bottom row with DICE score mentioned underneath. We show the segmentation results of our model with NCC and MSE along with SynthMorph and FourierNet.}
    \label{fig:supp_fig1}
\end{figure*}
\clearpage
\begin{figure*}[!htb]
    \centering

    \includegraphics[scale = 0.7]{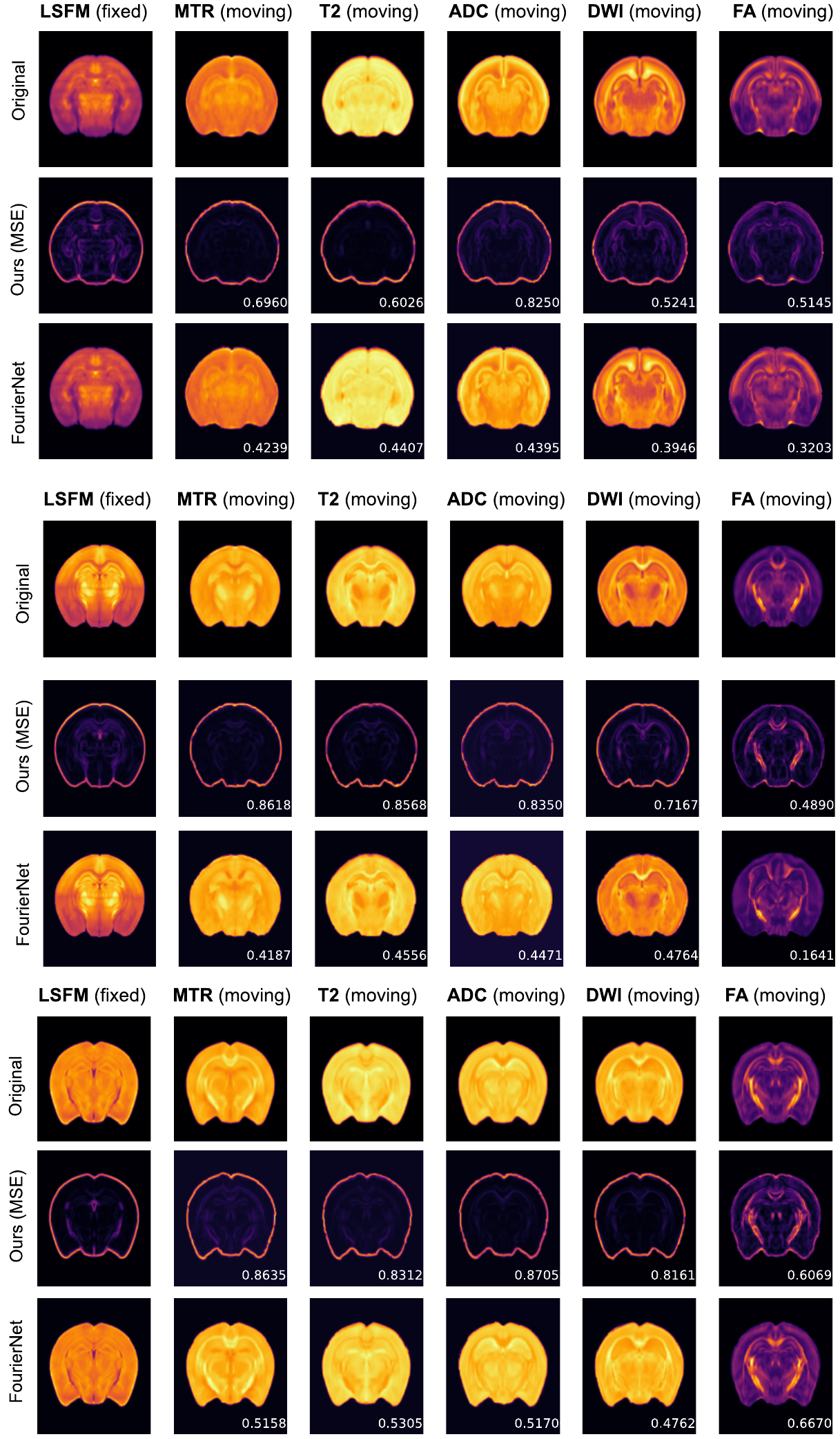}
    \caption{Qualitative result on DevCCF dataset's P04, P14 and P56 samples. The top row shows the original samples for each domain. The middle row shows the registration result of our model whereas the last row is the FourierNet's results. The DICE score between the aligned moving brain segmentation and the fixed segmentation is mentioned at the bottom right of each sample.}
    \label{fig:supp_fig7}
\end{figure*}

\end{document}